\RequirePackage{fix-cm}
\documentclass{svjour3}                     
\smartqed  
\usepackage{graphicx}
\usepackage{cite}
\usepackage{amsmath}
\usepackage{amsfonts}
\usepackage{color}
\usepackage{subcaption}
\DeclareMathOperator*{\argmin}{argmin}
\DeclareMathOperator*{\argmax}{argmax}

\begin{document}


\title{Retrieval and Registration of Long-Range Overlapping Frames for Scalable Mosaicking of In Vivo Fetoscopy}


\titlerunning{Scalable Mosaicking of In Vivo Fetoscopy}        

\author{Lo\"ic Peter \and Marcel Tella-Amo \and Dzhoshkun Ismail Shakir \and George Attilakos \and Ruwan Wimalasundera \and Jan Deprest \and S\'ebastien Ourselin \and Tom Vercauteren}


\institute{Lo\"ic Peter, Marcel Tella-Amo, Dzhoshkun Ismail Shakir, Jan Deprest, S\'ebastien Ourselin, Tom Vercauteren\at Wellcome / EPSRC Centre for Interventional and Surgical Sciences, University College London, London, United Kingdom
\and
Jan Deprest, Tom Vercauteren \at Centre for Surgical Technologies,  KU Leuven, Leuven, Belgium
\and
George Attilakos, Ruwan Wimalasundera \at University College Hospital, London, United Kingdom
}


\maketitle

\begin{abstract}

\textit{Purpose} The standard clinical treatment of Twin-to-Twin Transfusion Syndrome consists in the photo-coagulation of undesired anastomoses located on the placenta which are responsible to a blood transfer between the two twins. While being the standard of care procedure, fetoscopy suffers from a limited field-of-view of the placenta resulting in missed anastomoses. To facilitate the task of the clinician, building a global map of the placenta providing a larger overview of the vascular network is highly desired.

\textit{Methods} To overcome the challenging visual conditions inherent to in vivo sequences (low contrast, obstructions or presence of artifacts, among others), we propose the following contributions: (i) robust pairwise registration is achieved by aligning the orientation of the image gradients, and (ii) difficulties regarding long-range consistency (e.g. due to the presence of outliers) is tackled via a bag-of-word strategy, which identifies overlapping frames of the sequence to be registered regardless of their respective location in time.

\textit{Results}
In addition to visual difficulties, in vivo sequences are characterised by the intrinsic absence of gold standard. We present mosaics motivating qualitatively our methodological choices and demonstrating their promising aspect. We also demonstrate semi-quantitatively, via visual inspection of registration results, the efficacy of our registration approach in comparison to two standard baselines.

\textit{Conclusion}
This paper proposes the first approach for the construction of mosaics of placenta in in vivo fetoscopy sequences. Robustness to visual challenges during registration and long-range temporal consistency are proposed, offering first positive results on in vivo data for which standard mosaicking techniques are not applicable.

\end{abstract}

\section{Introduction}
\label{sec:intro}

Twin-to-twin transfusion syndrome (TTTS) is a condition affecting identical twin pregancies, where unexpected vascular anastomoses occur between two twins sharing a single placenta~\cite{baschat2011twin}. This results in a blood imbalance between the two twins. The current state-of-the-art curative procedure consists in laser photo-coagulation via fetoscopy of the abnormal vessel anastomoses located on the placenta. More precisely, surgeons perform a progressive visual exploration of the placenta, with the aim of localising and eliminating the anastomoses which allow a direct blood transfer between the two twins. Due to the difficulty of manipulating the fetoscope and due to the very limited field-of-view available at each timepoint to the surgeon, some anastomoses can be missed by the surgeon leading to an only incomplete treatment~\cite{lopriore2007residual}. To assist a clinician during TTTS surgery, mosaicking approaches are desirable to create a map of the placenta from a video acquired during fetoscopy. With the help of such a map, the field-of-view can be enlarged to facilitate the task of the clinician regarding the identification of yet unexplored areas of the placenta, and to provide a better overview of the topology of the vascular network.

Image mosaicking is a classical computer vision problem, where the panorama of a scene is built from a series of overlapping pictures. The most standard approach for stitching images consists of the registration of overlapping pairs via the detection and matching of landmarks~\cite{Brown2007}. Such feature-based methods have been successfully applied for some medical applications, such as retinal mosaicking~\cite{Prokopetc2017} and fibroscopic video mosaicking~\cite{Atasoy2008}. However, other type of clinical images may display a lack of texture, occlusions and other factors that make a landmark-based registration of a pair of images not reliable enough. To address this issue, alternative registration methods have been employed such as semi-dense registration method for dynamic view expansion in an in vivo porcine experiment~\cite{dense_surface_rec}. In~\cite{visual_odometry}, dense correspondences in an in vivo experiment were used, demonstrating improvements over RANSAC-based algorithms. Another example of dense registration was also succesfully applied for confocal microscopy~\cite{VERCAUTEREN2006media}. We refer to~\cite{review} for a more comprehensive review of the intersection between simultaneous localisation and mapping (SLAM), scene reconstruction and mosaicking in endoscopic procedures.

\begin{figure}
\centering
\includegraphics[scale=0.34]{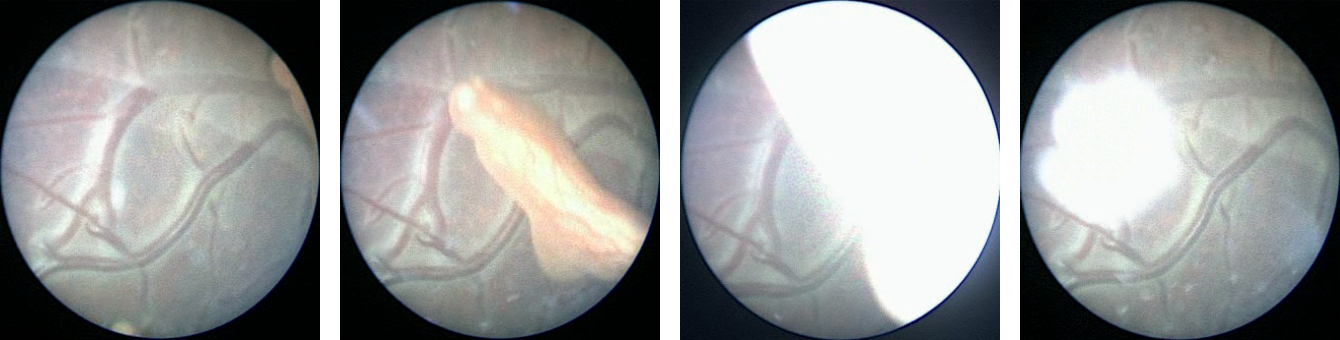}
\caption{\textbf{Visual challenges in in vivo fetoscopy.} Nearly consecutive frames of an in vivo sequence are shown. Together with a low contrast, in vivo data are subject to more or less severe occlusions due to fetal limbs or impurities present in the amniotic fluid.}
\label{fig:visual_challenges}       
\end{figure}

Closer to our application case, some works have attempted to perform mosaicking in  placental images. In~\cite{Reeff}, the authors report challenging situations that they tackle using a modified RANSAC algorithm. In~\cite{cnn_features}, a robust matching in phantom data was proposed via a new feature extractor algorithm using CNNs. External modalities such as 3D ultrasound~\cite{3d_ultrasound} or an electromagnetic tracker~\cite{marcel} were also investigated as means of guiding the mosaicking process. However, these approaches addressing the mosaicking of placenta images were until now limited to phantom and ex vivo data, for which visual properties are considerably different from the in vivo cases encountered in clinical scenarios. In the latter conditions, challenges such as repeated occlusions (e.g from fetal limbs or impurities present in the amniotic fluid, see Figure~\ref{fig:visual_challenges}) and low image contrast do not allow a successful application of standard landmark-based computer vision techniques.

In this paper, we propose a first approach towards the generation of placental mosaics from in vivo fetoscopy data. Our method combines (i) a registration method based on the alignment of gradient orientations, ensuring robustness to visual challenges inherent to in vivo acquisitions, and (ii) a strategy based on bags of visual words which allows the identification of pairs of overlapping frames located at arbitrary time points in the sequence. By retrieving and registering these key pairs of frames, the global consistency of the mosaic can be improved in a scalable manner. Qualitative results are reported and discussed based on real sequences and demonstrate first promising results towards the clinical use of mosaicking methods for TTTS surgery. In addition, we inspected visually the results of pairwise registration on an example sequence and labelled manually their quality, showing the benefit of our approach in comparison to two standard baselines: registration based on the robust matching of SURF-based keypoints, and dense image alignment based on normalised cross-correlation.

\section{An Approach for In Vivo Mosaicking}
\label{sec:methods}

\subsection{Problem Statement}
\label{sec:problem_statement}

Each frame of a fetoscopy sequence offers a partial view of the imaged placenta. Under the assumption that the placenta is planar, two arbitrary frames of the sequence are related by a homography transformation. Formally, given a sequence of $N$ frames $I_1, \ldots, I_N$ where each $I_i : \Omega_i \rightarrow \mathbb{R}^3$ is an RGB image defined over a domain $ \Omega_i \subset \mathbb{R}^2$, there exists for every pair of images $(I_i,I_j)$ a homographic warping $w_{j,i} : \Omega_i \rightarrow \Omega_j$ such that for every $\textbf{x} \in \Omega_i$, $I_i(\textbf{x})$ and $I_j(w_{j,i}(\textbf{x}))$ are visual measurements corresponding to the same location of the placenta. To create and visualise a mosaic, we propose the following approach. First, without loss of generality, the first frame of the sequence is defined as a reference frame located within the central part $\Omega_{\textrm{abs}}^1 \subset \Omega_{\textrm{abs}}$ of a (sufficiently large) mosaic image $M~:~\Omega_{\textrm{abs}}~\rightarrow~\mathbb{R}^3$. The mosaicking task aims at stitching together overlapping images to create a global map of the placenta or, in other words, at warping and placing each frame of the sequence on the corresponding part of the mosaic domain $\Omega_{\textrm{abs}}$. For every frame $I_i$, the corresponding subset $\Omega_{\textrm{abs}}^i$ of the absolute mosaic domain $\Omega_{\textrm{abs}}$ must be found. Equivalently, we propose to estimate a global homography $W_i$ such that $\Omega_{\textrm{abs}}^i = W_i(\Omega_{\textrm{abs}}^1)$. Note that $W_1$ is already defined as the identity via our choice of reference frame. Moreover, global and relative warpings are related via $W_i \circ W_j^{-1} = w_{i,j}$. To estimate the global warpings, we rely primarily on a series of pairwise registrations of overlapping frames which are directly conducted in the relative image domains $\Omega_i$. 

A fully sequential approach for mosaicking would register all consecutive frames and, with the obtained relative warpings $w_{k,k-1}$ for $2 \leq k \leq i$, compute the global warping $W_i$ of the $i$-th frame as follows:
\begin{equation}
W_i =  w_{i,i-1} \circ w_{i-1,i-2} \circ \ldots \circ w_{2,1}.
\label{eq:sequential_adjustment}
\end{equation}
If the estimation of the relative warpings is perfect, the equation above allows in theory a perfect mosaicking. However, in practice, errors in the estimation of each relative warping accumulate so that a clear mismatch can be observed when the fetoscope comes back to a previously visited location. This effect can even degenerate if the presence of occlusions makes the pairwise registration of two consecutive frames unfeasible, thereby breaking the chain of transformations~(\ref{eq:sequential_adjustment}). For increased robustness and temporal consistency over a large number of frames, it is therefore desired to register additional overlapping frames that are not necessarily consecutive (for example, the frames obtained when revisiting a portion of the placenta). Overall, if we denote $\mathcal{R}$ the set of couples of indices $(i,j)$ for which a registration has been performed and for which a resulting (possibly noisy) warping $\hat{w}_{i,j}$ estimating the true warping $w_{i,j}$ has thus been computed, and noticing that $W_i \circ W_j^{-1} = w_{i,j}$, we can look for global warpings $\hat{W}_1, \ldots, \hat{W}_N$ such that
\begin{equation}
\hat{W}_1, \ldots, \hat{W}_N = \argmin_{W_1, \ldots, W_N} \sum_{(i,j) \in \mathcal{R}} d(W_i \circ W_j^{-1} , \hat{w}_{i,j} ),
\label{eq:bundle_adjustment}
\end{equation}
where $d$ is a measure of dissimilarity between two warpings. The formulation (\ref{eq:bundle_adjustment}) is closely related to bundle adjustment and was proposed by Vercauteren et al. in the context of rigid transformations~\cite{VERCAUTEREN2006media}. We define the distance $d(w_1,w_2)$ between two warpings $w_1$ and $w_2$ defined over a rectangular image domain $\Omega$ as follows. First, we decide on a discrete set of reference points $\mathcal{X} = \lbrace\textbf{x}_1, \ldots, \textbf{x}_m\rbrace \subset \Omega$, which we choose in our case as a regular grid of step $3$ over $\Omega$. The distance $d(w_1,w_2)$ is then defined as
\begin{equation}
d(w_1,w_2) = \max_{\textbf{x} \in \mathcal{X}} \Vert w_1(\textbf{x}) - w_2(\textbf{x})  \Vert_2.
\label{eq:distance_warpings}
\end{equation}
This allows us to obtain an intuitive geometrical interpretation of the distance between warpings as the maximum deviation in terms of Euclidean distance over the set of reference points $\mathcal{X}$.

After having found estimates of the absolute warpings by solving (\ref{eq:bundle_adjustment}), a final mosaic can be created with blending algorithms~\cite{burt1983multiresolution,mahe2015motion}. In this paper, we focus on the accurate assessment of the global warpings, i.e. on the correct placement of the frames of the sequence on the mosaic. We used a standard publicly available technique~\cite{burt1983multiresolution} to generate the mosaics shown in this paper (Figure~\ref{fig:blend_mosaics}).

To summarise, we identified two crucial components for mosaicking:
\begin{enumerate}
\item Given two overlapping frames $I_i$ and $I_j$, we need a robust and reasonably fast way to register them to obtain a warping $w_{i,j}$.
\item To improve the consistency of the estimation over long timeframes,  it is crucial to identify additional overlapping frames that are not consecutive but located at timepoints arbitrarily far from another.
\end{enumerate}
We propose in this work a strategy to address separately these two challenging problems, respectively exposed in Section~\ref{sec:registration_gradients} and Section~\ref{sec:bag_of_words}, which takes into account the visual properties of in vivo sequences.

\subsection{Pairwise Registration of Consecutive Frames}
\label{sec:registration_gradients}

The traditional image stitiching technique~\cite{Brown2007} based on the detection and matching of landmarks (e.g. with a combination of SIFT and RANSAC) is prone to failure in in vivo sequences encountered in clinical conditions. The lack of constrast in the acquired images and the cluttered and varying aspect of the observed scene are challenges responsible for these difficulties. In this section, we present a registration method that addresses the pairwise registration of in vivo frames. Given the aforementioned challenges, we propose not to rely on landmarks. Instead, we perform a dense pixelwise alignment of the gradient orientations and propose a variant of the maximisation of the correlation of image gradients introduced by Tzimiropoulos et al.~\cite{Tzimiropoulos}, with two main differences exposed in details below. Aligning gradient orientations possesses several advantages: it is for example invariant to local changes of contrast and is suitable for registering accurately linear structures such as vessels, which matches the main clinical objective of mosaicking for TTTS surgery, i.e. the creation of an overview of the topology of the placental vascular network. Moreover, by focusing solely on gradient orientations and not on the gradient norms, each pixel is given the same weight, which naturally improves the robustness of the registration to visual artifacts and partial occlusions.

The registration task consists in estimating the true warping $w_{j,i}$ such that $I_i(\textbf{x})$ and $I_j(w_{j,i}(\textbf{x}))$ correspond to the same location for every $\textbf{x} \in \Omega_i$. With this formulation, $I_i$ is called the fixed image and $I_j$ the moving image. We parametrise the homographic warpings with a vector $\textbf{p} \in \mathbb{R}^8$ corresponding to the 8 coefficients of the canonical homographic representation, i.e. such that
\begin{equation}
w((x,y),\textbf{p}) = \left(\frac{p_1 x + p_2 y + p_3}{p_7 x + p_8 y + 1},\frac{p_4 x + p_5 y + p_6}{p_7 x + p_8 y + 1}\right).
\label{eq:homography}
\end{equation}
As discussed above, we propose to look for the registration warping $\hat{w}_{j,i}$ which aligns best the gradients of $I_i$ and $I_j$. Since we explicitly do not want to take into account the strength of the gradients, we first normalise the gradients of the fixed and moving images ensuring a unit gradient norm at every pixel. For a point $\textbf{x} \in \Omega_i$ of the domain of the fixed image, we denote $\Delta \theta(\textbf{x}, \textbf{p})$ the angle between the gradient of the fixed image and the gradient of the warped moving image at $\textbf{x}$. We define the final warping $\hat{w}_{j,i}$, i.e. the output of our registration method, as $\hat{w}_{j,i} = w(.,\hat{\textbf{p}})$ where
\begin{equation}
\hat{\textbf{p}} = \argmin_{\textbf{p}} \sum_{\textbf{x} \in \Omega_i} \sin^2\Delta \theta(\textbf{x}, \textbf{p}).
\label{eq:gradient_alignment_cost_function}
\end{equation}
Since $\sin^2t = \frac{1}{2} (1 - \cos 2t)$, the proposed approach can be seen as a variant of the maximisation of the correlation of image gradients~\cite{Tzimiropoulos} defined by
\begin{equation}
\hat{\textbf{p}} = \argmax_{\textbf{p}} \sum_{\textbf{x} \in \Omega_i} \cos \Delta \theta(\textbf{x}, \textbf{p}).
\label{eq:correlation_cost_function}
\end{equation}
We can identify two main differences between the two formulations. First, our pixelwise costs based on the sine function are minimal for $\Delta \theta = 0 $ or $\Delta \theta = \pi$, whereas the terms in (\ref{eq:correlation_cost_function}) are minimal for $\Delta \theta = 0$ only. Thereby, only the orientation (modulo $\pi$) of the gradients is taken into account in our cost function, and not their direction. It appears to be a useful property in practice: as we try to match vessels with an iterative method (see below), optimisation steps must be able to cross areas where gradient are oriented in opposite directions before reaching the optimal vessel alignment.
Having written our minimisation problem (\ref{eq:gradient_alignment_cost_function}) as a sum of squares, we are also able to use known results on non-linear least squares, and more precisely the forward additive version of the Lucas Kanade algorithm~\cite{Baker2004}. This formulation not only leads to simpler theoretical mathematical derivations, but also offers the possibility to use off-the-shelf optimised solvers for non linear least squares problem, such as the Ceres solver~\cite{ceres-solver} which includes classical optimisation techniques (the Gauss-Newton, Levenberg-Marquardt or Powell Dog-Leg algorithms, for example). 

\paragraph{Solving (\ref{eq:gradient_alignment_cost_function}) with the Gauss-Newton algorithm} To solve numerically the minimisation problem (\ref{eq:gradient_alignment_cost_function}), we use the fact that it is a non linear least squares problem to apply the Gauss-Newton algorithm, which, in the context of image registration, can also be seen as the forward additive version of the Lucas-Kanade algorithm~\cite{Baker2004}. To keep the following derivations as general as possible, we denote $N = \vert \Omega_i \vert$ and arbitrarily order the elements of $\Omega_i$ so that $\Omega_i = \lbrace \textbf{x}_1, \ldots, \textbf{x}_N \rbrace $, and we denote $M$ the number of parameters encoding the transformation, i.e. the size of the parameter vectors $\textbf{p}$. Applying the Gauss-Newton algorithm, we approximate iteratively the desired minimum with a series of parameters $\textbf{p}^{(1)}, \textbf{p}^{(2)}, \ldots$ such that
\begin{equation}
\textbf{p}^{(k+1)} = \textbf{p}^{(k)} - (\textbf{J}^T \textbf{J})^{-1} \textbf{J}^T \textbf{s} (\textbf{p}^{(k)}),
\label{eq:gauss_newton}
\end{equation}
where $\textbf{s} (\textbf{p}^{(k)})$ is the $N \times 1$ column vector $\left( \sin \Delta \theta(\textbf{x}_i, \textbf{p}^{(k)}) \right)_{1 \leq i \leq N}$ and $\textbf{J}$ is the $N \times M$ Jacobian matrix whose coefficients $J_{ij}$ are defined as
\begin{equation}
J_{ij} = \frac{\partial s_i(\textbf{p}^{(k)})}{\partial p_j} = \frac{\partial \Delta \theta(\textbf{x}_i, \textbf{p}^{(k)})}{\partial p_j} \cos \Delta \theta(\textbf{x}_i, \textbf{p}^{(k)}).
\label{eq:jacobian_definition}
\end{equation}
We denote $\textbf{g}_m(\textbf{x}_i, \textbf{p}^{(k)}) = \left( g_{m,x}(\textbf{x}_i, \textbf{p}^{(k)}), g_{m,y}(\textbf{x}_i, \textbf{p}^{(k)})\right) $ the gradient of the warped moving image at the location $\textbf{x}_i$, and $\theta_m(\textbf{x}_i, \textbf{p}^{(k)})$ (respectively $\theta_f(\textbf{x}_i))$ the angle of the gradient of the moving image (respectively the fixed image) at the location $\textbf{x}_i$. By definition, we have $\Delta \theta(\textbf{x}_i, \textbf{p}^{(k)}) = \theta_m(\textbf{x}_i, \textbf{p}^{(k)}) - \theta_f(\textbf{x}_i)$ and
\begin{equation}
\theta_m(\textbf{x}_i, \textbf{p}^{(k)}) = \arctan  \frac{g_{m,y}(\textbf{x}_i, \textbf{p}^{(k)})}{g_{m,x}(\textbf{x}_i, \textbf{p}^{(k)})},
\label{eq:arctan_angle}
\end{equation}
so that the coefficients of the Jacobian given in (\ref{eq:jacobian_definition}) can be written more explicitly as
\begin{equation}
J_{ij} = \frac{1}{\Vert \textbf{g}_m \Vert^2} \left( g_{m,x} \frac{\partial g_{m,y}}{\partial p_j} - g_{m,y} \frac{\partial g_{m,x}}{\partial p_j} \right) \cos \Delta \theta,
\label{eq:jacobian_expansion}
\end{equation}
where the dependencies in $\textbf{x}_i$ and $\textbf{p}^{(k)}$ were omitted for readability. We finally mention that, although the gradients $\textbf{g}_m(\textbf{x}_i, \textbf{p}^{(k)})$ could be computed at each iteration by warping the moving image and computing the gradient of the resulting warped image numerically, it is more efficient to precompute once for all the gradient of the (unwarped) moving image and obtain $\textbf{g}_m$ by warping this gradient and multiplying it by the Jacobian of the warping, by application of the chain rule~\cite{Tzimiropoulos}. The partial derivatives of $g_{m,x}$ and $g_{m,y}$ are obtained similarly.

\subsection{Ensuring Long-Range Consistency with Bag of Words}
\label{sec:bag_of_words}

As discussed in Section~\ref{sec:problem_statement}, evaluating the set of global homographies from a series of pairwise registrations of consecutive frames inevitably leads to an accumulation of registration errors. In the most extreme case, the chain of transformation can even be broken if a registration is not feasible at all, for example in the presence of a full occlusion. However, fetoscopy conditions naturally lead to long sequences during which the surgeon follows vessels one by one, resulting in the presence of numerous overlapping areas in the sequence. Therefore, introducing additional constraints from the registration of non temporally consecutive frames may provide the redundancy to compensate for the drift and the robustness to failed registrations of consecutive frames. If we could have a reliable way to decide from the registration result if the registration was successful (see Section~\ref{sec:registration_validity}), we could in theory register all image pairs to extract the highest amount of information. However, registering all image pairs is computationally intractable for long sequences and would probably introduce more redundancy than required. Therefore, we need an efficient way to predict, from their visual appearance, the pairs that are worth registering.

\begin{figure}
\centering
\includegraphics[scale=0.3]{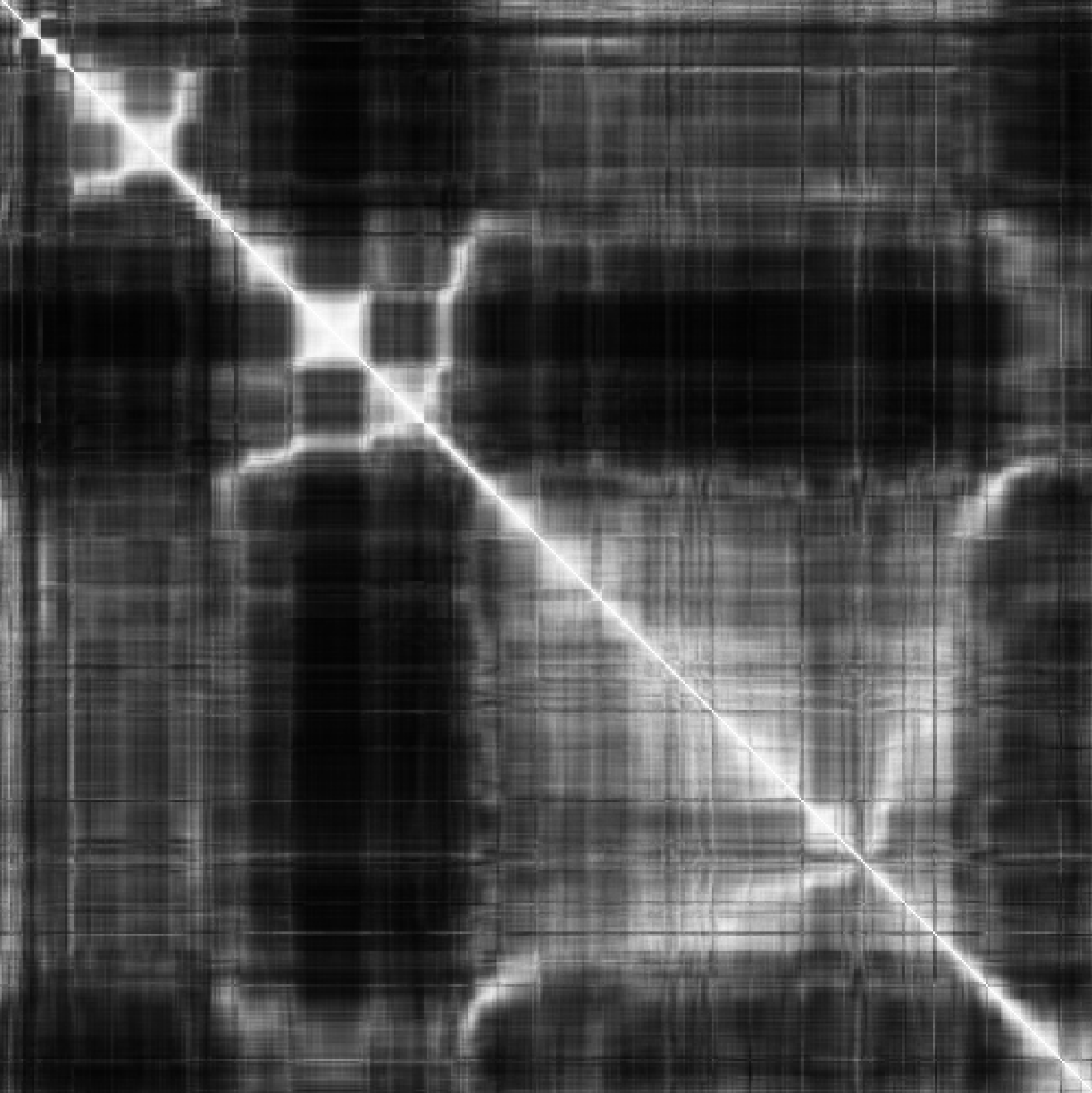}
\caption{\textbf{Similarity matrix.} Every entry $(i,j)$ of this matrix states how visually similar the frames $I_i$ and $I_j$ are. Note how the camera following a vessel back and forth creates branches that are orthogonal to the diagonal line.}
\label{fig:similarity_matrix}       
\end{figure}

Following an idea introduced in computer vision~\cite{Ho2007}, we adopt a strategy based on bags of visual words~\cite{Csurka04visualcategorization} to efficiently identify frames sharing a similar content without the need to register them. We sample dense keypoints using the VGG descriptor~\cite{Simonyan} and perform a $K$-means clustering over the full video to obtain a vocabulary of $K$ visual words. Each image is then described by a signature vector $\textbf{v} \in \mathbb{N}^K $ encoding the frequence of each visual word in the image. The visual similarity between two images $I_i$ and $I_j$ is then computed as the cosine distance between the two associated signature vectors $\textbf{u}$ and $\textbf{v}$, i.e
\begin{equation}
s(I_i,I_j) = \frac{\sum_{k=1}^K \textbf{u}_k \textbf{v}_k}{\sqrt{\sum_{k=1}^K \textbf{u}^2_k} \sqrt{\sum_{k=1}^K\textbf{v}^2_k}}.
\label{eq:cosine_similarity}
\end{equation}
By computing this similarity measure for every pair of images in the videos, we obtain a similarity matrix on which the revisiting of previous locations is apparent (Figure~\ref{fig:similarity_matrix}). The construction of this matrix is more scalable than attempting the registration of all pairs and, in fact, only requires approximate nearest neighbours for which algorithms in linear time (e.g. FLANN) are available. Figure~\ref{fig:similarity_matrix} shows an example of similarity matrix. In this example video, the trajectory followed by the clinician is "star-shaped": every vessel is followed until its extremity, before following it back until the last intersection, usually at the coord insertion site. The timepoints and patterns corresponding to these "back and forth" trajectories are apparent on the similarity matrix as lines orthogonal to the diagonal.

To define the set of additional candidate registrations to be included in the bundle adjustment formulation (\ref{eq:bundle_adjustment}), we simply rely on a threshold on the similarity above which the registration of a pair is tried. The choice of this threshold is mainly driven by computational considerations, as it is directly related to the number of attempted registrations which are going to be performed before solving the bundle adjustment problem.

\subsection{Assessing the Validity of a Registration}
\label{sec:registration_validity}

In the previous subsections, we described a robust method to register two images, as well as a way to retrieve pairs of non-consecutive images which share a visual overlap so that a registration of these pairs can be attempted, in addition to the consecutive pairs. Each registration is used as a term in the bundle adjustment formulation (\ref{eq:bundle_adjustment}) and acts as a constraint in the estimation of the global homographies necessary to create the mosaic. However, in practice, the obtained registrations are not always accurate, due for example to the registration optimiser being trapped in a local minimum, to a failure of the retrieval of overlapping frames (leading to the unfeasible registration of two frames for which there is no overlap), or to the presence of a large occlusion in a frame caused for example by fetal limbs. Assessing the correct registrations within the attempted ones is of critical practical importance to filter out these wrong constraints added in the bundle adjustment, which would bias the final estimation of global homographies. 

In this work, we declare a registration as successful if it is close enough from the identity (in the sense of the distance $d$ defined in Section~\ref{sec:problem_statement}), and if the gradient-based cost function which was minimised in~(\ref{eq:gradient_alignment_cost_function}) is sufficiently small in comparison to the costs obtained with random warpings sampled around the identity. Although this empirical strategy proved to be effective in practice, we plan to investigate as future work more sophisticated methods, such as consistency checks over cycles of frames~\cite{datteri2015validation} or learning-based approaches.

%
%
%
%
%
%
%
%
%
%
%
%
%

\section{Experimental Validation}

\subsection{Implementation Details}

We implemented our method in C++ using the OpenCV library. The bundle adjustment minimisation problem (\ref{eq:bundle_adjustment}) was solved numerically using the Levenberg-Marquardt algorithm with the Ceres solver~\cite{ceres-solver}. By restricting the warpings to affine transformations, i.e. homographies where $p_7 = 0$ and $p_8 = 0$ with the notations introduced in (\ref{eq:homography}), convergence to a visually sound solution was achieved, even when starting from identity warpings as initialisation. If one desired to work with general homographies instead, closed-form solutions could be used to obtain initialisations close enough from the global optimum~\cite{schroeder2011wacv}. The pairwise registrations using the forward additive Lucas Kanade algorithm were performed in a Gaussian pyramidal fashion with $6$ levels (where each level is a blurred and scaled down version of the original image, as implemented in OpenCV), starting by registering the images at a coarse level and refining progressively the warpings by increasing the resolution. In the case where a registration is rejected at a level of the pyramid, it was reinitialised as the identity for the next level. For increased robustness, we performed for each pair both a forward and backward registration (i.e. switched fixed and moving image) and kept the registration leading to the lowest cost. The bags of visual words were computed using the OpenCV implementation, with default parameters in the extraction of the VGG descriptors.

The pairwise registration between two frames takes approximately $1$ second. These registrations remain the main bottleneck in practice: on our example video of $600$ frames, solving the bundle adjustment problem takes a few minutes, as does the construction of the similarity matrix from the bag of words. Therefore, there is a direct linear relationship between the total computational time and the amount of attempted registrations of retrieved pairs of frames. For a given computational budget, this time can be controlled by registering a predefined amount of pairs with the maximum similarity, as mentioned at the end of Section~\ref{sec:bag_of_words}. Given our current implementation and set of parameters, the total mosaicking pipeline took about $3$ hours.

\subsection{Qualitative Results}
\label{sec:qualitative_results}

\begin{figure}
\centering
\includegraphics[scale=0.55]{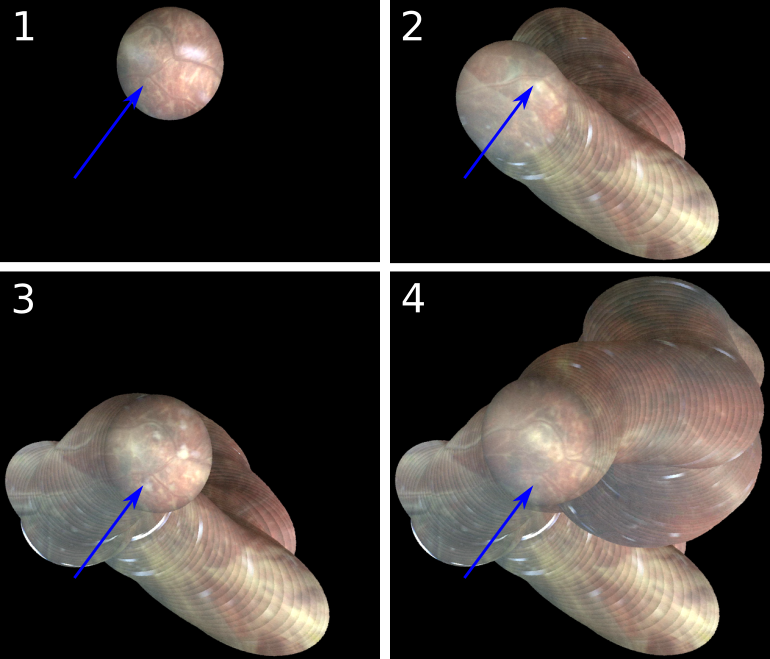}
\caption{\textbf{Retrieval and registration of overlapping frames for long-range consistency.} By retrieving and successfully registering frames showing the same vascular intersection (marked with blue arrows), the location of this area in the mosaic is kept stable, ensuring an improved global consistency.}
\label{fig:results_bow}       
\end{figure}

Due to the nature of in vivo acquisitions, a ground truth for placenta mosaicking is not available. This makes a quantitative evaluation of our approach very difficult. Nevertheless, we demonstrate first promising visual results with our approach, which we discuss qualitatively in this section. Figure~\ref{fig:results_bow} shows, on an example video of $600$ frames, the appearance of the mosaic obtained after the estimation of the global homography of each video frame. To facilitate the visual interpretation of the results and relate it more easily to the actual content of our example video, we limit the display of the mosaic to truncated versions of the video at different timepoints (from $1$ to $4$, chronologically). Each frame of the video is pasted chronologically onto the mosaic according to the global homography obtained after the offline optimisation. Note that the global optimisation has been run once for all on the whole video, and that these chronological timepoints are only introduced for the visualisation of our results. Figure~\ref{fig:results_bow} illustrates the global consistency of the mosaic: although an area showing a ~$\rotatebox[origin=c]{-90}{Y}$~-shaped vascular intersection is visited $4$ times during the sequence (once at each chosen snapshot), this intersection is correctly placed at the same location in the mosaic over time. This is due to the fact that video frames containing this intersection have been recognised as similar in terms of visual content and successfully registered, adding additional constraints in the global optimisation which lead to improved temporal consistency.

\begin{figure}[t]
    \centering
    \begin{subfigure}[b]{0.49\textwidth}
    \centering
        \includegraphics[width=\textwidth]{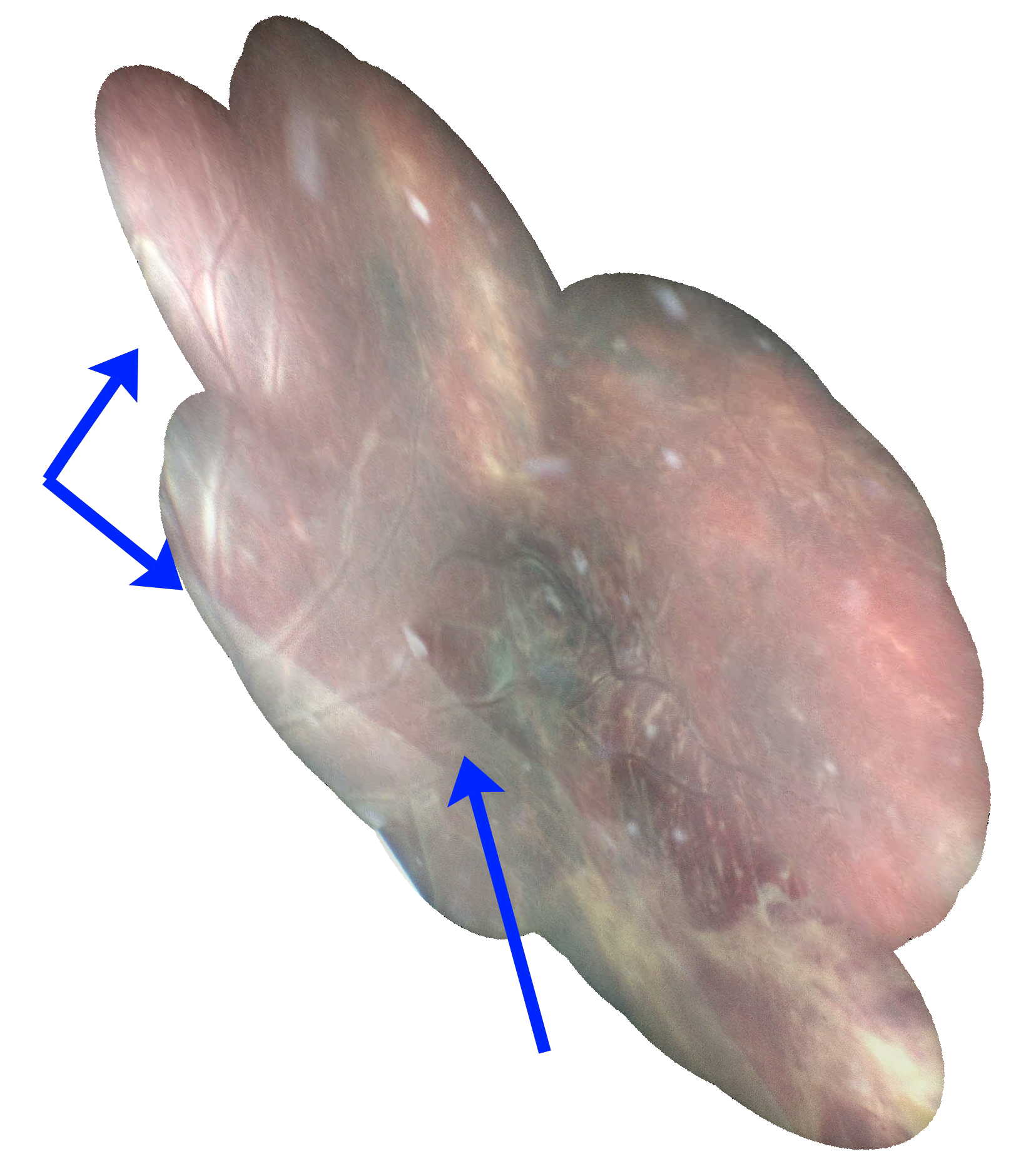}
        \caption{Purely sequential alignments}
        \label{fig:blend_drift}
    \end{subfigure}
    \begin{subfigure}[b]{0.49\textwidth}
    \centering
        \includegraphics[width=\textwidth]{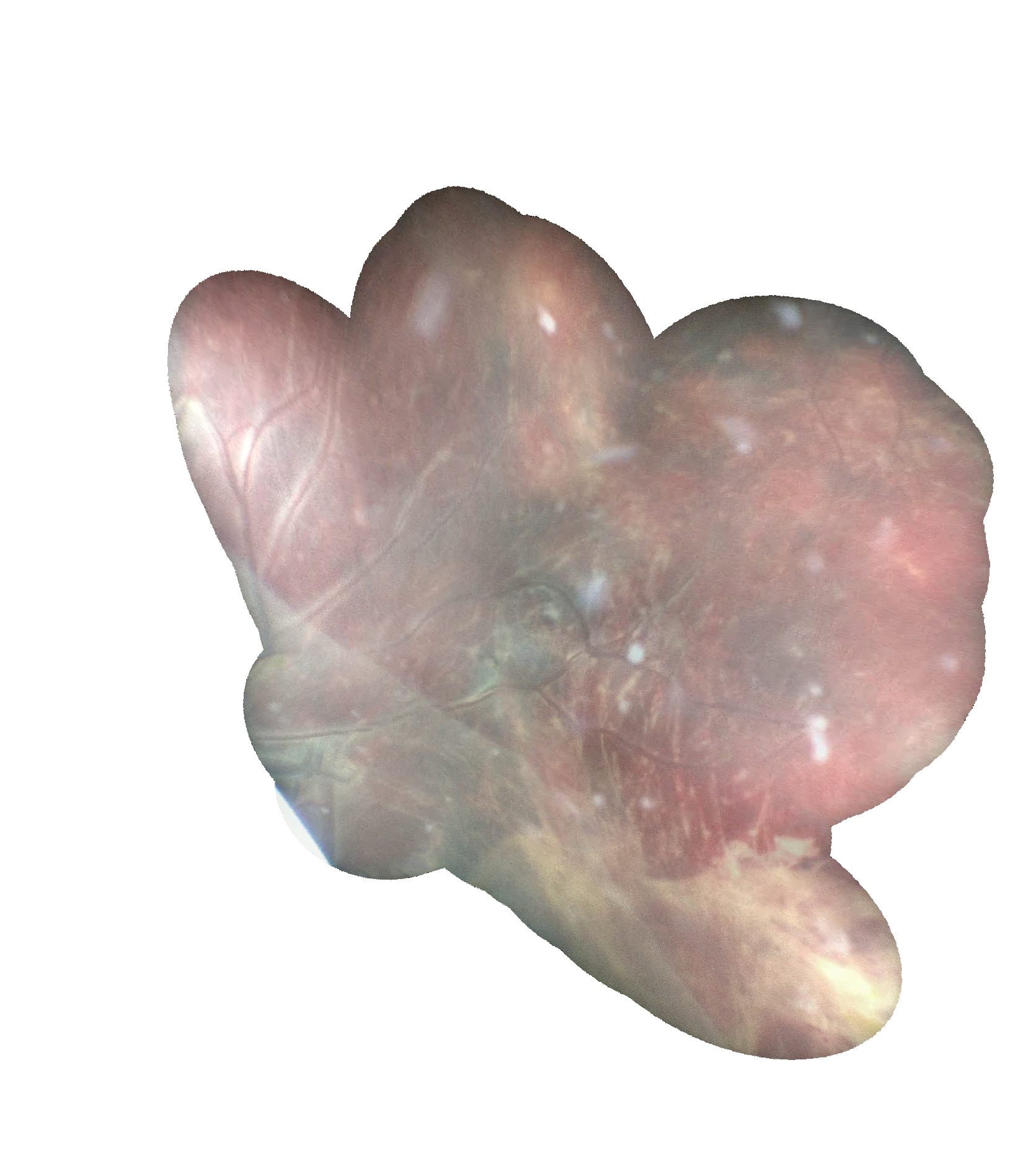}
        \caption{Our approach}
         \label{fig:blend_bow}
    \end{subfigure}    
    \caption{\textbf{Mosaics obtained after blending.} We show two example mosaics obtained with and without the introduction of long-range consistency. Without long-range consistency (Figure~\ref{fig:blend_drift}), an accumulation of errors between pairwise registrations occurs, so that misalignments are caused when revisiting locations (such as the vessels on the top left part or the membrane on the bottom left part, both marked with blue arrows). 
    } 
    \label{fig:blend_mosaics}
\end{figure}

We show on Figure~\ref{fig:blend_mosaics} the resulting mosaics, where frames are merged in a seemless fashion following the method of Burt and Adelson~\cite{burt1983multiresolution} publicly available in the software Enblend\footnote{http://enblend.sourceforge.net/}, where one frame every $5$ frames is used for blending. We show the mosaics obtained when our gradient-based registration method was used, with and without the long-range consistency with the bag of words formulation (respectively Figure~\ref{fig:blend_bow} and Figure~\ref{fig:blend_drift}). In the purely sequential case (Figure~\ref{fig:blend_drift}), the drifting behavior can be seen via the generally more distorted aspect of the mosaic, as well as clear misalignments (best seen when compared to Figure~\ref{fig:results_bow}), for example of the vessels on the top left part and of the membrane of the amniotic sac (linear demarcation on the bottom left part).

\subsection{Manual Evaluation of Pairwise Registrations}

In addition to the qualitative results discussed in Section~\ref{sec:qualitative_results}, we compare our gradient-based registration approach with two baselines. Our first baseline is a classical image stitching technique used in computer vision available in OpenCV which consists in detecting SURF-based keypoints in each image and align the images with RANSAC. SURF features were chosen over SIFT as they performed slightly better empirically, confirming the observations made by Reeff et al. on their own placenta images~\cite{Reeff}. Our second baseline is obtained by replacing our gradient-based similarity measure by the normalised cross-correlation, keeping the rest of our Gauss-Newton optimisation framework unchanged.

To conduct the comparison, we considered our example video of $600$ frames and assessed visually the quality of the 599 sequential pairwise registrations for each baseline. Each registration was manually labelled as either \textit{correct}, \textit{incorrect} or \textit{doubtful} (for ambiguous cases where the correctness of the registration is difficult to assess visually). To remove any subjective bias, the $1797$ registrations to be labelled were randomly shuffled so that the annotation was done without knowledge of the approach which was actually evaluated in each case. Table~\ref{tab:quantitative_registrations} summarises the count of registration belonging to each category for the three methods, and confirms the benefit of our approach based on the alignment of gradient orientations.

\begin{table}[h]
\center
\begin{tabular}{|c| c c c |}
		\hline
		 Method & Correct registrations & Doubtful & Incorrect registrations \\
		\hline
		\texttt{SURF-RANSAC} & 56 (9.3\%) & 83 (13.9\%) & 460 (76.8\%) \\
		\texttt{NCC} & 29 (4.8\%) & 61 (10.2\%) & 509 (85.0\%) \\
		\texttt{Ours} & 477 (79.6\%) & 67 (11.2\%) & 55 (9.2\%) \\
		\hline
\end{tabular}
\caption{Evaluation of pairwise registrations}
\label{tab:quantitative_registrations}

\end{table}


\section{Conclusion}

We proposed a first step towards the mosaicking of placenta images from in vivo fetoscopy data. A robust registration method based on the alignment of gradient orientations addresses the visual challenges inherent to in vivo sequences which prevent the successful application of classical mosaicking techniques used in computer vision. Moreover, the global consistency of the mosaics were improved via a retrieval strategy based on bags of visual words, which identifies pairs of frames which are worth registering, regardless of their respective location in time. Qualitative results were shown and discussed to illustrate the relevance of our approach.

%
%

\begin{acknowledgements}
\textbf{Funding:} This work was supported by  Wellcome / EPSRC [203145Z/
\\16/Z; NS/
A000050/1; WT101957; NS/A000027/1] and EPSRC [EP/L016478/1]. Jan Deprest is supported by the Great Ormond Street Hospital Charity. This work was undertaken at UCL and UCLH, which receive a proportion of funding from the DoH NIHR UCLH BRC funding scheme.
\\
\textbf{Conflict of Interest:} The authors declare that they have no conflict of interest.
\\
\textbf{Ethical approval:} All procedures performed in studies involving human participants were in accordance with the ethical standards of the institutional and/or national research committee and with the 1964 Helsinki declaration and its later amendments or comparable ethical standards.
\\
\textbf{Informed consent:} Informed consent was obtained from all individual participants included in the study.

\end{acknowledgements}

\bibliographystyle{plain}
\bibliography{Mosaicking}   


\end{document}